\documentclass{article}

\usepackage{times}
\usepackage{graphicx} % more modern
\usepackage{subfig} 

\usepackage{natbib}

\usepackage{algorithm}
\usepackage{algorithmic}

\usepackage{hyperref}

\usepackage[accepted]{icml2014}

\icmltitlerunning{Deep Learning for Class-Generic Object Detection}

\begin{document} 

\twocolumn[
%AC:  Brody:  feel free to change title.  Just trying to put into submittable state.
\icmltitle{Deep learning for class-generic object detection}

% unless you've provided the [accepted] option to the icml2014
\icmlauthor{Brody Huval}{brodyh@stanford.edu}
%\icmladdress{Your Fantastic Institute,
%            314159 Pi St., Palo Alto, CA 94306 USA}
\icmlauthor{Adam Coates}{acoates@stanford.edu}
\icmlauthor{Andrew Ng}{ang@stanford.edu}
%\icmladdress{Their Fantastic Institute,
%            27182 Exp St., Toronto, ON M6H 2T1 CANADA}
%\icmladdress{Stanford University Computer Science Dept., 353 Serra Mall, Stanford, CA 94305 USA}

\icmlkeywords{Object Detection, Objectness, Class-generic, Deep Learning}

\vskip 0.3in
]

\begin{abstract} 
We investigate the use of deep neural networks for the novel task of class-generic object detection.  We show that neural networks originally designed for image recognition can be trained to detect objects within images, regardless of their class, including objects for which no bounding box labels have been provided.  In addition, we show that bounding box labels yield a $1\%$ performance increase on the ImageNet recognition challenge.
\end{abstract} 

\section{Introduction}

The task of separating objects from background is fundamental for many computer vision tasks. This has led to much research on localizing and classifying objects by using object segmentation, object detection, and region proposals. Currently, most detectors are trained individually for each object class, which requires a class label and a bounding box for all images. Unfortunately, in this approach it is difficult to transfer information from previously trained detectors to novel classes where bounding box labels may not be available. This situation is common in current datasets, which often have many class labels but incomplete bounding box labels. In this work, we aim to overcome these challenges by training separately from bounding box labels and class labels, enabling our system to learn even when only one of these labels is available. This approach harnesses the notion of object-ness \cite{endres2010category, alexe2012measuring,uijlings2013selective} to build a deep neural network \cite{krizhevsky2012imagenet} able to detect novel objects where bounding box labels have not been provided.
%AC:  made minor changes to last couple sentences.. what do you think?

%Current strategy is to train detectors for each object class, though this requires
%many images annotated with the location and class of the object.
%Unfortunately, this approach doesn't allow us to transfer the learned detection abilities
%to novel object classes where we have few or no bounding box labels. 

%%AC:
% re-state the problem:
%  explain current detection strategy [DPM], etc.
%  challenging because of labels;  datasets sizes;  [mention VOC/ImageNet label counts?]
%  more generally, would like to learn to detect objects without knowing class and 
% avoid the problem of needing class labels and box labels for every object.

%AC:
One successful approach to object detection is to train a single detector for each class of objects (for example, the Deformable Parts Model (DPM)~\cite{felzenszwalb2010object}).  In this approach, one discriminatively trains a set of detectors on each individual class. This strategy generally has proven useful on the Pascal VOC detection challenge due to the limited number of classes, each of which includes many bounding box labels. In other cases, however, where we may have an abundance of class labels, but few or no bounding box labels, it is not clear how to apply this same strategy.  For example, the Image-Net dataset has 14 million class labels but only about $7\%$ are labeled with bounding boxes~\cite{deng2009imagenet}.

Recently, region proposal algorithms have shown good performance in object detection pipelines by proposing class-generic locations for further classification~\cite{endres2010category, alexe2012measuring,girshick2013rich,uijlings2013selective}. They attempt to measure object-ness within an image by training on all bounding boxes labels, regardless of class, in hopes of building a single detector for all classes. 

% propose an alternative
While training from only bounding box labels potentially enables a detector to locate novel classes never seen before, it may perform poorly due to having too few training examples and failing to exploit the wealth of class labels available in datasets like Image-Net. We propose to train a detector to localize objects while also exploiting object class labels by separating the recognition and detection problems. We show that by pretraining our detector on class labels and then on object locations, we can increase its performance in detecting previously seen objects, while nearly retaining its ability to localize objects for which we have no bounding box labels.

\section{Related works}
\cite{szegedy2013deep} have used a similar neural network for object detection in Pascal VOC. Like our approach, they avoided the use of sliding windows or region proposals, and instead directly used a deep neural network for predicting object locations. However, their work focuses on only a handful of classes from Pascal VOC, and five different networks are trained for each class. In contrast, we train a single network able to provide class-generic object detections. % which is pre-trained for image recognition

Region proposal algorithms are typically shallow methods that focus on high-recall object-ness detection~\cite{endres2010category, alexe2012measuring,uijlings2013selective}. Therefore, they return hundreds to thousands of potential bounding boxes for evaluation, which is still a large reduction over the number of evaluations required for the sliding window approach. These potential locations are then input to a stronger classification algorithm such as a deep Convolutional Neural Network (CNN) to identify the object.  In our work, we focus on high-precision detection, using non-max supression to reduce our predictions to a set of likely detected objects.

%Due to multiple tasks we use to increase performance, image recognition \& object detection, our work also has similarities to ~\cite{collobert2008unified}.

\section{Object Detection from Neural Networks}
\subsection{Model}
The Convolutional Neural Network (CNN) we use is similar to that proposed by~\cite{krizhevsky2012imagenet}  for object classification. The network consists of five layers of convolution followed by two densely connected layers. Every layer applies a Rectified Linear Unit (ReLU) as its non-linearity. Only the first, second, and fifth layers use Local Contrast Normalization (LCN) and max pooling. The final output is a $4096$ dimensional feature vector for the image.  See~\cite{krizhevsky2012imagenet} for details.

\subsection{Bounding Box Training}
In the image classification results from~\cite{krizhevsky2012imagenet}, the final feature vector is input to a softmax layer which provides a probability distribution over class labels. In our work, we instead use a softmax layer to provide a probability distribution over a discretized space of bounding boxes.  This 4-dimensional space encodes the x-y position, the scale, and the aspect ratio of a bounding box.

Because the softmax layer treats every output label independently, the network will receive the same loss for bounding boxes with high or low overlap with the ground truth, which is not ideal. This also leads to low counts of each label during training.  To resolve this, instead of placing a one or zero at each label, we place a Gaussian distribution centered at the correct label. The result is a smaller loss when a bounding box similar to the ground truth is predicted. To allow multiple bounding boxes in each image, a Gaussian distribution is placed at each location and the labels are re-normalized to sum to one.  During evaluation, multiple boxes are predicted by applying non-max suppression to the resulting probability distribution over bounding boxes.

\subsection{Classification pretraining}
When training a CNN for image recognition, the network is learning discriminative filters that help in detecting the different classes, while ignoring potentially distracting generic backgrounds.  As a result, the task of recognition and detection are related, and information from one task can improve results on the other.  To implement this intuition, we optionally pretrain our network using the image recognition task.  Our results will confirm the usefulness of this intuition:  Pretraining on image recognition turns out to increase performance on class-generic object detection.  Conversely, pretraining on object detection can increase image recognition performance. 
% AC:  learning discriminative filters that help in detecting the $1000$ different synsets in the challenge --- 
% maybe "training a function able to identify 1000 classes from the ImageNet
% dataset"   In general:  try to avoid extra detail that is not important
% for understanding the key idea of the paragraph.  (Also, "synsets" is sort
% of a form of "jargon" that many will not understand.)

\section{Experiments}
To perform our experiments, we use multiple GPUs for training~\cite{coates2013deep, krizhevsky2012imagenet}.

% AC: To perform our experiments, we use multiple GPUs for training~[cite coates2013,krizhevsky2012, others..?]

\subsection{Dataset}
% describe why we chose Imagenet 2012 challenge over Imagenet 2013 detection and Pascal VOC

All experiments were performed on the Imagenet 2012 Localization Challenge~\cite{deng2009imagenet}. This dataset provides $1.2$ million classification images with $592,000$ including bounding box labels over $1000$ different categories/synsets. 

\subsection{Evaluation}

\begin{figure}[h!]
  \centering
    \includegraphics[width=0.5\textwidth]{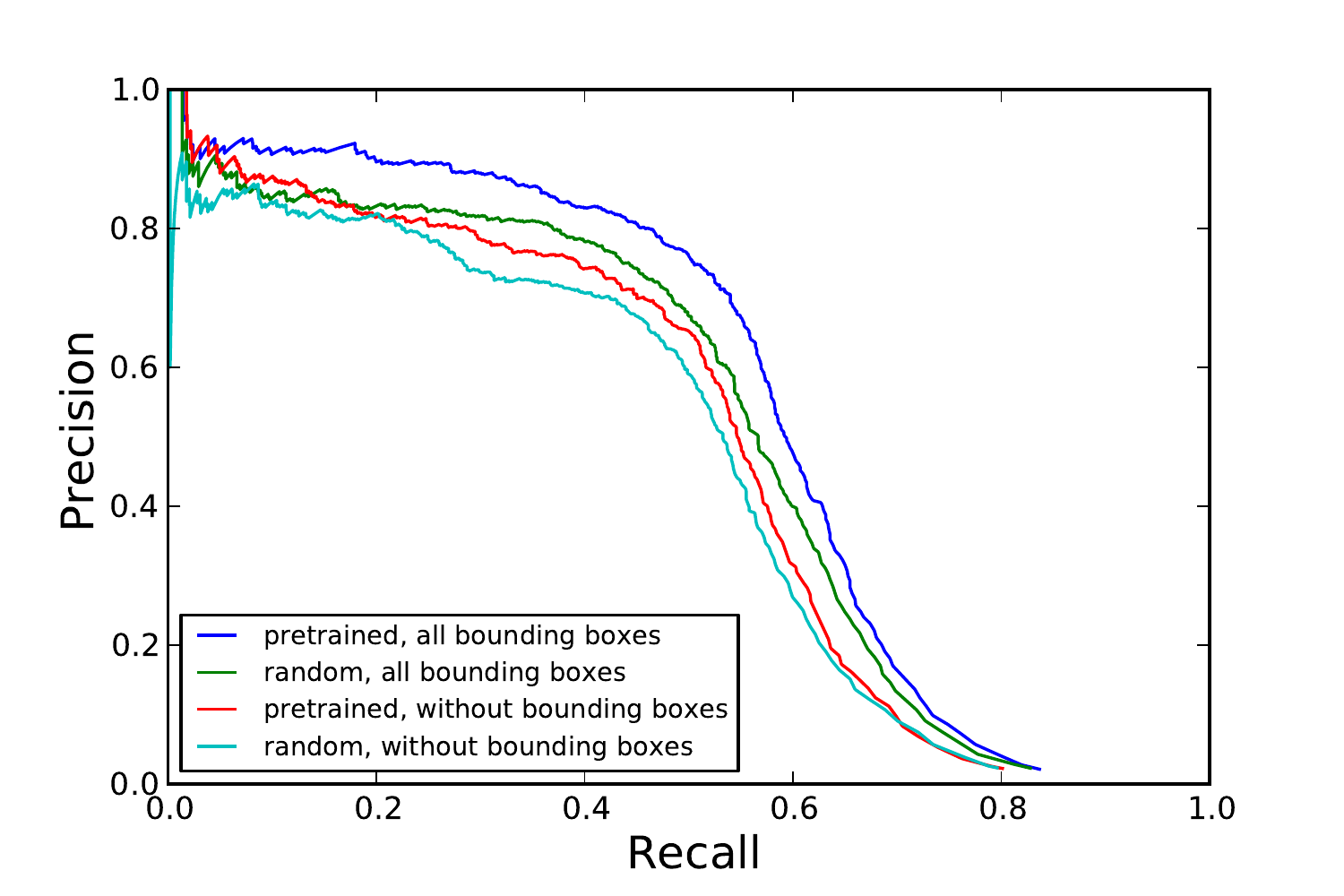}
%\vskip -0.1in
\caption{Precision Recall curves.}
\label{fig:precision_recall}
%\vskip -0.1in
\end{figure}

\begin{figure*}[th!]
\centering
\subfloat[~]{\label{fig:f1}\includegraphics[width=1.1in]{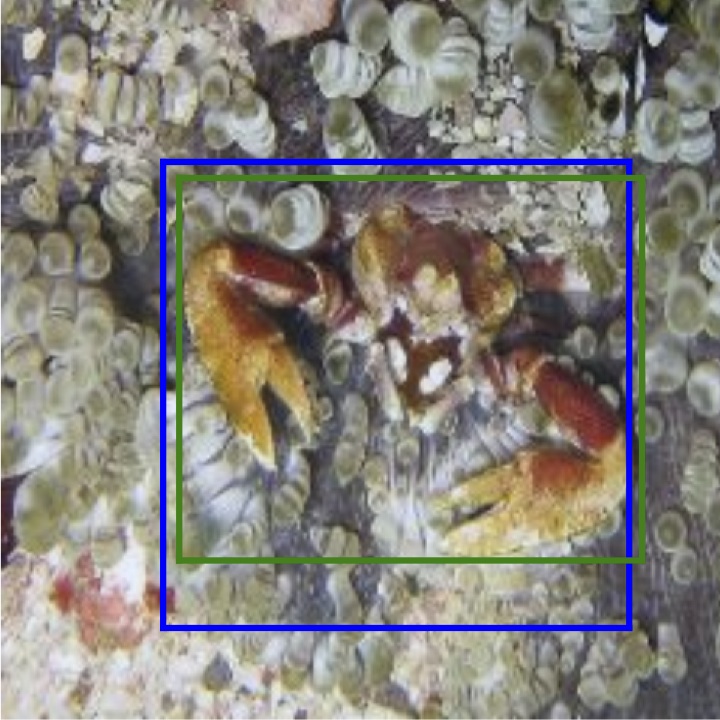}}
    \subfloat[~]{\label{fig:f2}\includegraphics[width=1.1in]{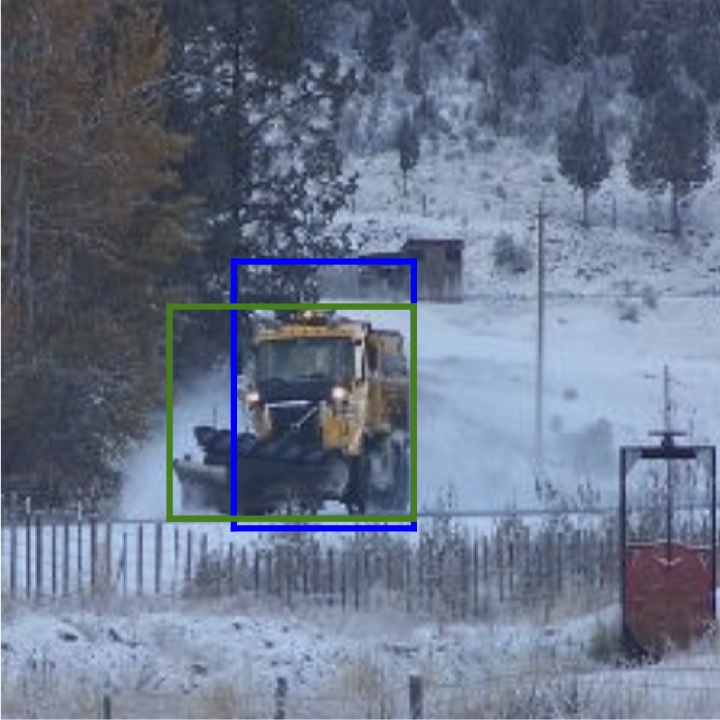}}
    \subfloat[~]{\label{fig:f3}\includegraphics[width=1.1in]{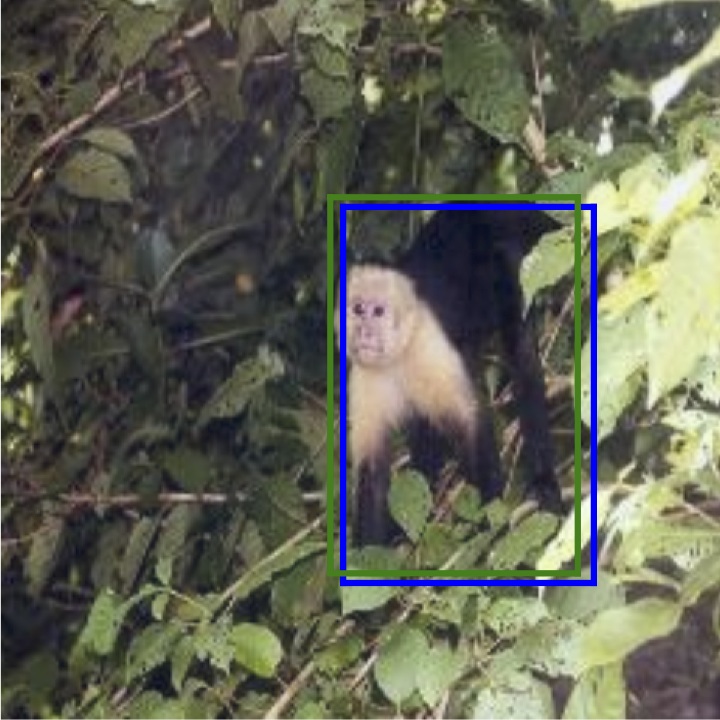}}
    \subfloat[~]{\label{fig:f4}\includegraphics[width=1.1in]{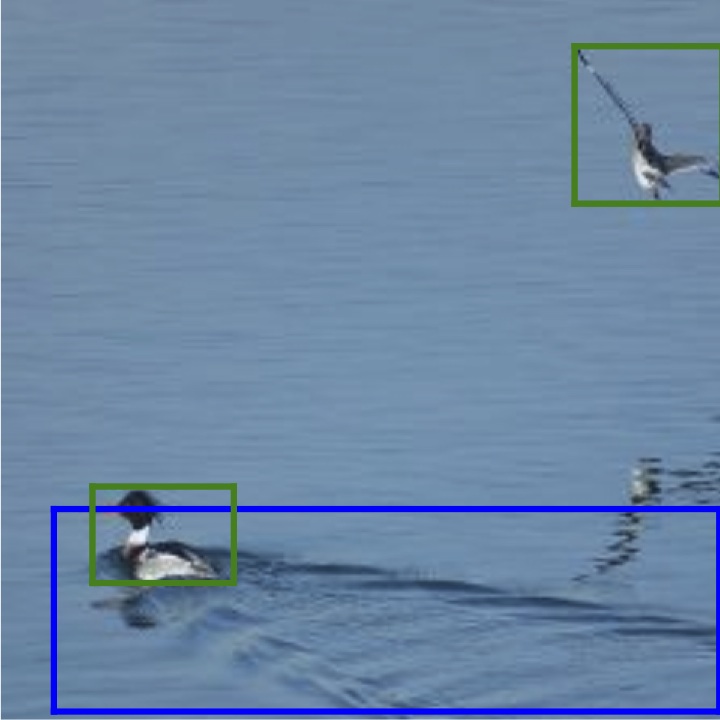}}
    \subfloat[~]{\label{fig:f5}\includegraphics[width=1.1in]{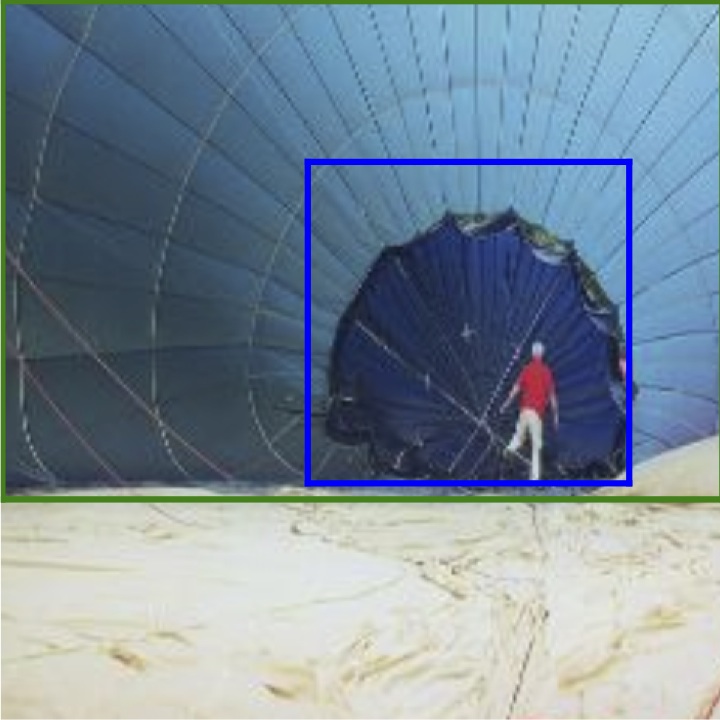}}
    \subfloat[~]{\label{fig:f6}\includegraphics[width=1.1in]{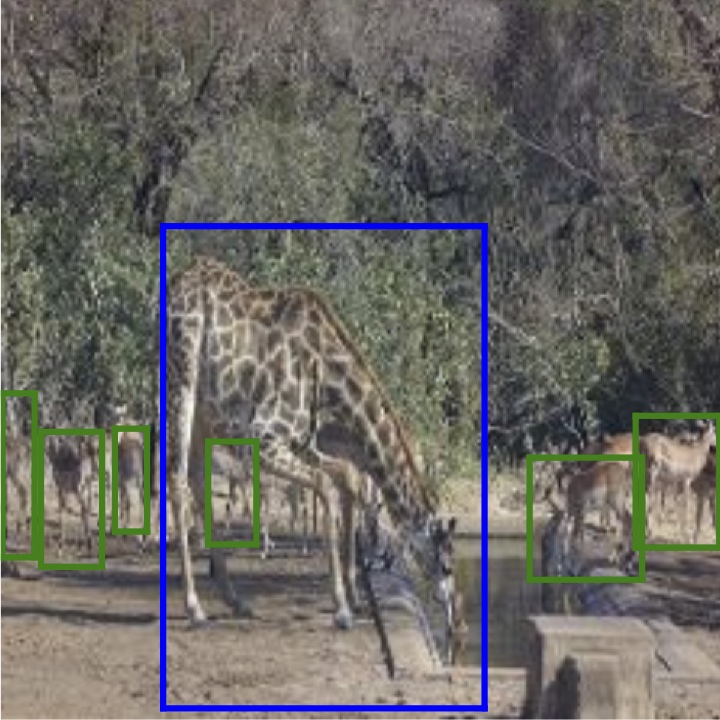}}
%  \vskip -0.1
  \caption{(a)-(c) Correct (d)-(f) Incorrect detections for classes without bounding box labels. True labels shown in green, predictions in blue.}
%  \vskip -0.1
  \label{fig:detections}
\end{figure*}

To show that our network is capable of detecting objects, despite never having seen their bounding boxes, we randomly chose $100$ object classes out of the $1000$ ImageNet Challenge classes and train without their bounding boxes.  We then evaluate the performance on a validation set only containing the $100$ object classes held out during training.  This was performed with one network starting from random weights, and another starting from a network pretrained on image recognition for all $1000$ object classes.  Our results are shown in Table~\ref{tab:aucs}.  To measure the drop in performance from not having the bounding boxes from $100$ classes available, the performance for both networks, pretrained and random, are reported when trained on all $1000$ bounding box classes, shown in Table~\ref{tab:aucs_all}. Precision recall curves are also shown for these four cases in Figure~\ref{fig:precision_recall}. Examples of correct and incorrect detections from our network on held out bounding box classes are shown in Figure~\ref{fig:detections}.

%This experiment was performed with pretraining on all $1000$ image classification labels, and with starting from random weights. The precision recall curves are shown in Figure~\ref{fig:precision_recall} and the corresponding F1 scores for each case are shown in Table~\ref{tab:f1scores}. Some exmples of correct and incorrect detections from our network on unseen bounding box labels are shown in Figure~\ref{fig:detections}.
%AC:  "on unseen bounding box labels" --> on novel classes without bounding box labels.  ???

\begin{table}
    \begin{center}
    \caption{AUC after training without $100$ classes bounding box classes. }
    \begin{tabular}{| l | c | c |}
    \hline & AUC \\ \hline
        pretrained & 0.475 \\ \hline
        random & 0.448 \\ \hline
    \end{tabular}
    \label{tab:aucs}
\end{center}
\end{table} 

\begin{table}
    \begin{center}
    \caption{AUC after training on all bounding box labels.}
    \begin{tabular}{| l | c | c |}
    \hline & AUC \\ \hline
        pretrained &  0.545 \\ \hline
        random & 0.498 \\ \hline
    \end{tabular}
    \label{tab:aucs_all}
\end{center}
\end{table} 

Finally, we looked at the impact of training to detect objects (without the corresponding class labels) on image recognition performance.  We found that by initializing an image recognition network with the weights from a trained object detection network, we reduce our top-5 error by $1\%$. %from $23.76\%$ to $22.75\%$
%AC:  We suspect that further improvements might be obtained with additional tasks or training on both tasks
% simultaneously~\cite{collobert2008unified}.

\section{Conclusion}

%AC:  advice on conclusions:   restate what your paper was about, and what
% you showed;  this is your last chance to just pound one clear idea and make them remember it.  Maybe 1 sentence at most about future work, etc.  [I personally don't like future work bits;  they tend to sound hand-wavy and mostly get ignored.  Might be fine for workshop paper though.]

By using a single deep neural network we have investigated a method for object-ness detection, capable of exploiting both class and bounding box labels. Our network is able to generalize to classes for which it has never seen bounding box labels while benefitting from class labels when available.  In addition, we have also found that the object-ness detection task yields a modest improvement in the ImageNet recognition challenge, which does not involve detection.

%In future work, we hope to use our object detector as a region proposal algorithm, which are used in the current state-of-the-art object detection pipelines~\cite{girshick2013rich}. % AC:  is this salient?

% BH: no we can take it out
%  AC: haha.

%This work can also be seen as a move towards a universal image representation. One which is able to encode not only class but location of an object within an image.

%We are optimistic about scaling this algorithm up to train on all Imagenet classification and bounding box data. This could then be used for a much higher precision region proposal algorithm, which are a key component to the current state of the art on Pascal VOC ~\cite{girshick2013rich}.

\bibliography{iclr2014_workshop}
\bibliographystyle{icml2014}

\end{document}